\documentclass[sigconf]{acmart}

\acmConference[SPLC'18]{22nd International Conference on Software Product Line}{10--14 September, 2018}{Gothenburg, Sweden}

\usepackage[T1]{fontenc} 
\usepackage{amsmath}
\usepackage{amssymb}

\usepackage{multirow}
\usepackage{multicol}
\usepackage{minibox}
\usepackage{todonotes}
\usepackage{listings}

\usepackage{hyperref}
\usepackage{graphicx}
\setcopyright{none}

\usepackage{subfig}

\usepackage{algorithm}
\usepackage{algorithmic}

\usepackage{tcolorbox}

\usepackage{xspace}
\newcommand\etal[0]{\emph{et al.}\xspace}
\newcommand{\ie}{\emph{i.e.,}\xspace}
\newcommand{\eg}{\emph{e.g.,}\xspace}

\newcommand{\etc}{\emph{etc.}\xspace}
\newcommand{\wrt}{w.r.t.\xspace}

\newcommand{\td}[1]{}

\usepackage{stmaryrd}

\newif\ifdraft\drafttrue

\ifdraft
\newcommand\todos[1]{\todo[inline]{TODO (all): #1}}
\newcommand\ma[1]{\todo[color=green!40,inline]{TODO (Mathieu): #1}}

\newcommand\jmj[1]{\todo[color=yellow!40,inline]{TODO (JMJ): #1}}
\newcommand\rb[1]{\todo[color=blue!35,inline, caption={}]{TODO (Razieh): #1}}

\newcommand\lnb[1]{\todo[inline,color=red!20!blue!10,bordercolor=red!20!blue!40]{TODO (Leo): #1}}

\else
\newcommand\todos[1]{}
\newcommand\ma[1]{}

\newcommand\jmj[1]{}
\newcommand\rb[1]{}

\newcommand\lnb[1]{}
\fi

\begin{document}

\title{Towards Adversarial Configurations for Software Product Lines}

\author{Paul Temple}
\affiliation{%
  \institution{Univ Rennes, IRISA, Inria, CNRS}
  \city{Rennes}
  \country{France}
}

\author{Mathieu Acher}
\affiliation{%
  \institution{Univ Rennes, IRISA, Inria, CNRS}
  \city{Rennes}
  \country{France}
}

\author{Battista Biggio}
\affiliation{%
  \institution{University of Cagliari}
  \city{Cagliari}
  \country{Italy}
}

\author{Jean-Marc J\'{e}z\'{e}quel}
\affiliation{%
  \institution{Univ Rennes, IRISA, Inria, CNRS}
  \city{Rennes}
  \country{France}
}

\author{Fabio Roli}
\affiliation{%
  \institution{University of Cagliari}
  \city{Cagliari}
  \country{Italy}
}

\begin{abstract}
Ensuring that all supposedly valid configurations of a software product line (SPL) lead to well-formed and acceptable products is challenging since it is most of the time impractical to enumerate and test all individual products of an SPL. 
 Machine learning classifiers have been recently used to predict the acceptability of products associated with unseen configurations. 
 For some configurations, a tiny change in their feature values can make them pass from acceptable to non-acceptable regarding users' requirements and vice-versa.
In this paper, we introduce the idea of leveraging these specific configurations and their positions in the feature space to improve the classifier and therefore the engineering of an SPL.
Starting from a variability model, we propose to use Adversarial Machine Learning techniques to create new, adversarial configurations out of already known configurations by modifying their feature values.
Using an industrial video generator 
we show how adversarial configurations can improve not only the classifier, but also the variability model, the variability implementation, and the testing oracle.   


\end{abstract}

\settopmatter{printfolios=true}

\maketitle

\section{Introduction}

\emph{Software product lines (SPLs)} promise to deliver custom products out of users' configurations. 
 Based on their specific needs, users select some configuration options (or features) that are combined at the implementation level for eventually deriving a desired software product. 
 Real-world SPLs offer hundreds to thousands of configuration options through runtime parameters, conditional compilation directives, configuration files, or plugins~\cite{pohl-etal2005,apel2013book}. 
 
The abundance of options can be seen as a standing goal of an SPL, but it also challenges the engineering of SPLs.
In particular, the configuration process can be tedious as combinations of options might not be functionally valid or lead to unacceptable performances (\eg execution time) for a given product. 
 It is extremely painful for an end-user, an integrator or a software developer to discover, late in the process (at compilation time or even worse at exploitation 
 time), that her carefully chosen set of options is actually invalid or non-acceptable (for whatever definition of acceptability).
 
 In fact, ensuring that all supposedly valid configurations of an SPL lead to well-formed and acceptable products has been a challenge for decades~\cite{batory2007,metzger2007,thuem2014}. For instance, formal methods (such as model checking) and static program analysis have been developed~\cite{DBLP:conf/pldi/BoddenTRBBM13,DBLP:journals/fac/StruberRACTP18,classen2010,classen2011,DBLP:journals/jlp/BeekFGM16,DBLP:conf/icse/NadiBKC14}. 
 Dynamic testing is another widely used alternative as it is sometimes the only way to reason about functional and quantitative properties of product of an SPL. However, it is most of the time impractical to enumerate, measure, and test all individual products of an SPL. 
 As a result, \emph{machine learning (ML)} techniques are more and more considered to predict the behavior of an SPL out of a (small) sample of configurations~\cite{SGKA:ESECFSE15,guo2015,guo2013,DBLP:conf/isola/BeekFGS16,siegmund2013,DBLP:conf/sigsoft/OhBMS17}. 
 In particular, ML classification techniques can be used to predict the acceptability of unseen configurations -- without actually deriving the variants~\cite{DBLP:journals/software/TempleAJB17,temple:hal-01323446}.
 A central problem then remains: the statistical ML algorithm can produce classification errors (by construction). Non-acceptable variants may still be generated out of supposedly valid configurations; or invalid configurations actually correspond to acceptable video variants.

Our intuition is that errors (if any) come from the proximity, in the feature space, between non-valid and valid configurations.
This proximity may intertwine both acceptable and non-acceptable configurations, increasing the complexity of ML functions separating the two configurations classes (acceptable or non-acceptable). Prediction errors come from the fact that ML techniques are estimators and might not be able to cope with this complexity.
Our idea is to exploit the estimator to find "blind spots" in the separating function and target these particular areas to better define the separating function.
It will improve the ML classifier associated to an SPL and, thus, better capture the space of valid and acceptable configurations.
%

We propose to use \emph{Adversarial ML (AdvML)} techniques which automatically create configurations that are specifically designed to lie in areas of the configuration space where the confidence in the ML decision is low (\ie close to the boundary and thus where valid and non-valid configurations are close). To our knowledge, no works have used AdvML to reach the goal of improving an SPL.


The contributions of this paper can be summarized as follows:
\emph{i)} we motivate the problem with an industrial video generator;
\emph{ii)} we describe a conceptual framework for SPL engineering in which ML classifiers are central; 
\emph{iii)} we introduce the idea of adversarial configurations and detail how AdvML techniques are suited for generating them; 
\emph{iv)} we show that adversarial configurations can help to improve not only the classifier of an SPL, but also the variability model, the variability implementation, and the testing oracle of an SPL (\eg an industrial video generator). 
\section{Motivating Case Study}
\label{sec:video_gen}



We consider a representative SPL of our problem, an industrial video generator called MOTIV (more details can be found in~\cite{galindoISSTA2014,temple:hal-01323446,mauricio2018}). 
The goal of MOTIV is to produce synthetic videos out of a high-level, textual specification; such videos are then used to benchmark Computer Vision based systems under various conditions. Variability management is crucial to produce a diverse yet realistic set of video variants, in a controlled and automated way.

MOTIV is composed of a variability model that documents possible values of 80 configuration options. Each option has an impact of the visual characteristics on generated videos. There are Boolean options, categorical (an enumeration) options (\eg for including fog or blur in a scene) and real-value options (\eg to deal with the amount of dynamic or static noise). 
To realize variability, the MOTIV generator relies on Lua code that takes as input a configuration file and produces a custom video file (see Figure~\ref{fig:framework}). 
A highly challenging problem of MOTIV is that, out of the $\approx10^{100}$ possible configurations, some of the corresponding videos are not acceptable.
For example, there is too much noise or blur in some of them; or additional objects, such as trees, obstruct the view to the scene making the processing too difficult and unrealistic, \etc 

To overcome this limitation, our early attempt was to rely on ML classification techniques to predict the acceptability of unseen video variants. 
We used an automated procedure (\ie a testing oracle) to compute and determine whether visual properties of a video were acceptable. 
Learned constraints were extracted and injected into the variability model to reduce its configuration space.
%
 Despite good accuracy and interpretable results, errors remain -- supposedly valid configurations may still generate non-acceptable video variants or invalid configurations actually correspond to acceptable video variants. 
Thus, it is important to be sure that the classifier does not have "blind spots" and that the decisions it makes are as close as possible from the decision that the oracle would make.
Said differently, we want to reduce the number of errors made by the classifier with regards to the oracle.




\section{SPL and Machine Learning Classifier}

The MOTIV case study is an instance of a more general problem. We propose a conceptual SPL framework to describe the problem and its entities. In particular, we show the central role of ML classifiers.

\label{sec:framework}

\subsection{Basic SPL framework}

\begin{figure}
\centering
\includegraphics[scale=0.21]{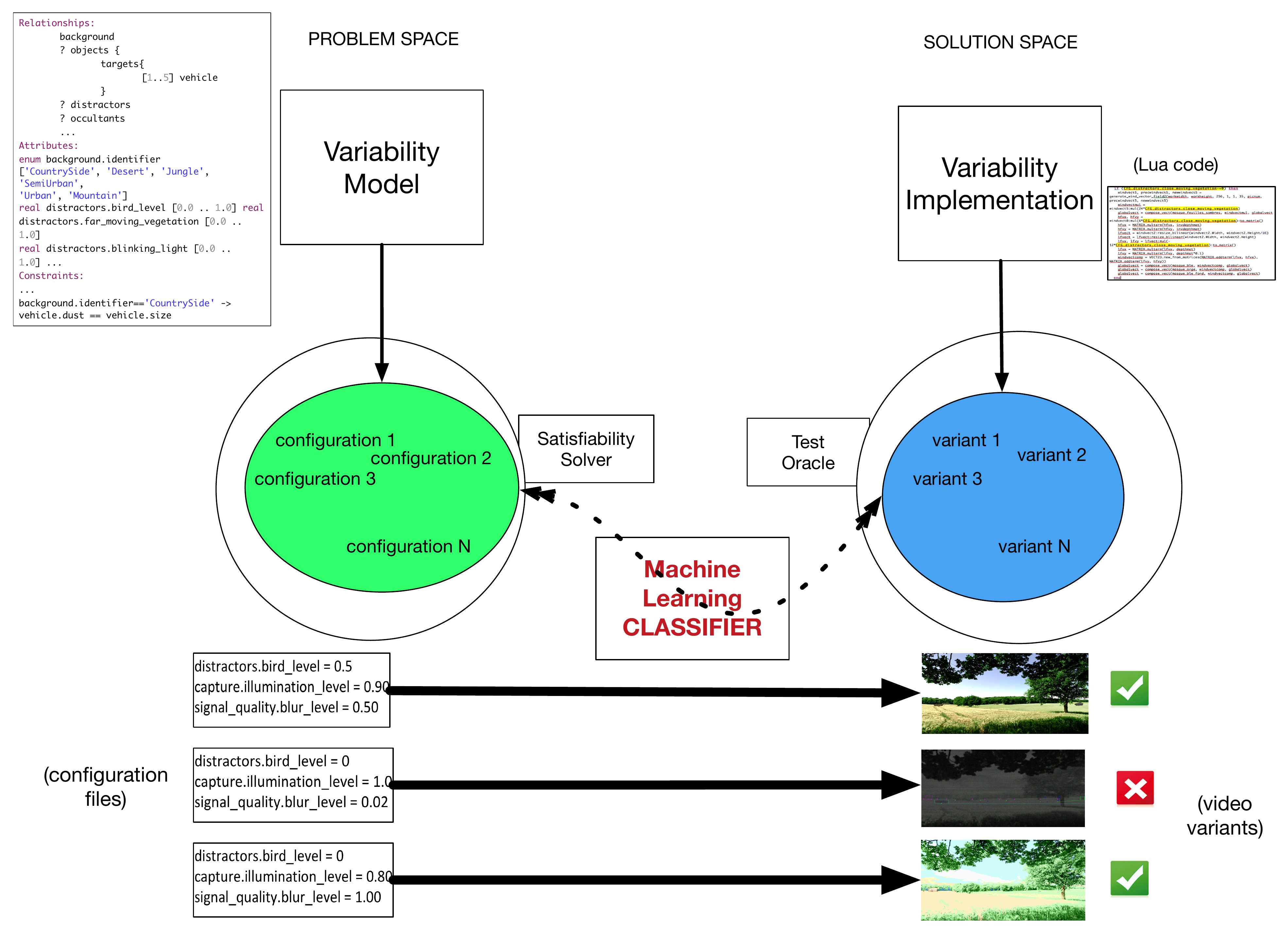}
\caption{Software product line and ML classifier}
\vspace*{-6mm}
\label{fig:framework}
\end{figure}

Figure~\ref{fig:framework} depicts the different entities of the framework as we illustrate them on the MOTIV case.

\textbf{Variability modeling.} A \emph{variability model} defines the configuration options of an SPL; various formalisms (\eg attributed feature models, decision models) can be employed to structure and encode information~\cite{berger2013,benavides2010}.
A variability model typically defines domain values over each configuration options allowing to bound values that they can take. 
 Moreover, as not all combinations of values are permitted, it is common to write additional constraints between options (\eg mutual exclusions between two Boolean options). 


A \emph{configuration} is an assignment of values to every individual options.
Because of constraints and domain values, the notions of \emph{valid} and \emph{invalid} configurations emerge.
That is, some values and combinations of configuration options are accepted while others are rejected.
A \emph{satisfiability solver} (\eg SAT, CSP, or SMT solver) is usually employed to check the validity of configurations and reason about the configuration space of a variability model. 
Such a reasoning procedure is usually sound and complete. 


\textbf{Variability implementation.}
Configurations are only an abstract representation of a variant (or a product); there is a need to shift from the problem space to the solution space and concretely realize the corresponding variants with actual code.
Different implementation techniques can be used such as \#\textit{ifdefs} which give instructions to the compiler or runtime parameters given to programs. In the case of MOTIV, the Lua generator uses different parameters to execute a given configuration and produce a variant. 
 

In some cases, configurations can lead to undesirable variants despite being valid.
For instance, in the case of MOTIV, some video variants contained too much noise.
The \emph{test oracle} is a procedure to determine whether a variant is acceptable or not. 
 In Figure~\ref{fig:framework}, the oracle gives a label (green/acceptable or red/non-acceptable).
As the number of variants can be large, it is desirable, as often as possible, to automate the procedure (\eg with Unit test cases). 
In the case of MOTIV, we can hardly ask a human to visually assess all possible video variants. We implemented a C++ procedure for computing visual properties of a video. 
%
 If a variant is considered non-acceptable by the oracle, then, there is a difference between the decision given by the solver within the problem space and the testing oracle within the solution space.
Problems can occur in the transformation from problem space to solution space:
in particular, the code can be buggy; the oracle can be hard to automate and thus introduce approximations and/or errors; the variability model might miss some constraints.

\subsection{Machine Learning (ML) Classifier}

So far, the conceptual framework presented in Figure~\ref{fig:framework} is rather traditional.
We now describe the role of ML classifiers.


\textbf{Why ML classifiers are needed?}
A typical problem from SPL engineering is to ensure the integrity between the problem space and the solution space.
That is, all valid configurations of the variability model in the problem space must be associated to an acceptable variant in the solution space.
 In the case of MOTIV, determining whether a video variant is acceptable can only be done after the derivation and through dynamic testing. Furthermore, it is most of the time impossible to execute all configurations to assess whether corresponding variants are acceptable or not.
Beyond MOTIV, many SPLs are in this situation: the configuration space is huge and dynamic testing can only be done over a small sample. 
 ML techniques are precisely here to generalize observations made over known configurations to never-seen-before configurations. 

\textbf{ML classifiers to predict the label of unseen configurations.}

From a formal point of view: we consider a classification algorithm $f: X \mapsto Y$ that assigns samples represented in a feature space $x \in X$ to a label in the set of predefined classes $y \in Y$.
In MOTIV, only two classes are defined in $Y = \{ -1, +1 \}$: acceptable and non-acceptable.
The classifier $f$ is trained on a dataset $D$ sampled from a variability model and constituted of a set of pairs ($x_i$, $y_i$) of configurations defined in $X$ and associated labels. 
The classifier builds a separating function that can be later used to predict the class of previously unseen configurations represented in the feature space $X$.
 A testing oracle is used to compute and associate labels to configurations in the training sample. When presenting unseen configurations, a ML classifier will hopefully predict correct labels without actually deriving corresponding variants.  


\begin{figure}
\centering
\includegraphics[scale=0.5]{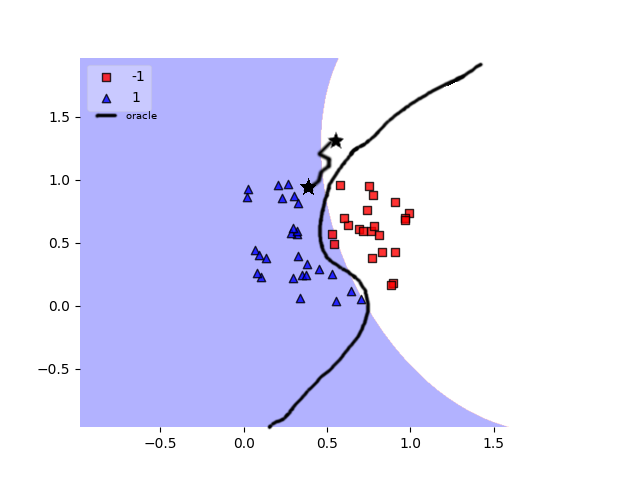}
\vspace*{-4mm}
\caption{Adversarial configurations (stars) are at the limit of the separating function learned by the ML classifier}
\label{fig:classif_errors}
\vspace*{-4mm}
\end{figure}

Unfortunately, the separation can make prediction errors since the classifier is based on statistical approaches and a (small) training sample. %
 Figure~\ref{fig:classif_errors} illustrates a set of configurations (triangles and squares) in a 2D space (\ie one dimension represents one configuration option) that is used to learn a separation (shown as the transition from the blue/left to the white/right area).
The solid black line represents the target oracle that the classifier is supposed to fit.
We can clearly see that the built separation is an approximation of the target oracle as:
going away from the center of the image, the two functions diverge;
two squares are already misclassified as being triangles.
The algorithm approximates as it tries to mitigate the complexity of the function \wrt the number of errors it makes.
\section{Using Adversarial ML}
\label{sec:advML}



\textbf{Principles.} Our goal is to reduce the number of errors performed by the classifier.
A first, simple method is to pick random configurations and, whenever a divergence happens between the decision made by the classifier and the oracle, the configuration is added to the training set with the label given by the oracle and build a new classifier.
Going further: after a random configuration have been picked, 
the configuration is transformed into its associated variant. Tests on this variant are executed and finally the oracle can decide.
In the end, because the configuration was chosen randomly, the classifier cannot be exploited at all.
Its predictions must be check in order to know if it disagrees with the oracle to further retrain it.

Instead of randomly choosing configurations, we would like to guide their generation to increase chances of obtaining a classification error. 
Getting back to Figure~\ref{fig:classif_errors}, the notion of confidence in the prediction emerges as the two classification errors are made. As the two squares lie close to the separation and knowing that they are misclassified (thanks to their real labels), the classifier will give a low confidence in its decision.
This piece of information can be exploited to guide the generation of new configurations towards similar areas where the confidence in the prediction is low.

This is somehow similar to Generative Adversarial Nets (GANs) idea~\cite{gan2014} 
as a classifier and adversarial configurations interact.
However, instead of a generative model, we propose to use other AdvML techniques in order to find "blind spots" in the classifier (\ie poorly explored areas in the feature space; areas with few known configurations that are possibly far from each other).
Specifically targeting such areas in order to create new configurations and include them into the training set will rise the confidence given into predictions leading to fewer prediction errors.

\textbf{Using evasion attack.} Biggio \etal~\cite{biggio2013evasion} present an AdvML technique, called evasion attack, that can be used after a classifier is trained.
Regarding some classifier implementations, confidence values might not be directly retrieved and an estimate of the confidence response has to be calculated.
Then, a gradient descent algorithm is used \wrt this estimation of the response in order to target directly area of low confidence.
Finally, attack points (\ie copy of actual known configurations) are modified following the gradient direction and become new configurations.

Adding these new configurations to the training set of the classifier and retrain it might help constrain the space of acceptable variants as the classifier will perform fewer prediction errors and will generally give a higher confidence in its predictions.
It is important to recall that this method does not choose randomly configurations and it only exploits the classifier after it has been trained that does not involve the use of the oracle.
Getting back to Figure~\ref{fig:classif_errors}, using an adversarial attack might create points lying around the points distribution (\ie central part of the image).
For instance, we might create adversarial configurations on the top and/or bottom parts of the image, in areas where we can observe strong divergence between the classifier and the oracle.
The principle is shown by the $\star$ configurations (representing initial and final positions of an adversarial configuration).
Taking these new configurations into account by retraining the classifier, we force the reduction of the gap between the two functions, in turn, reducing prediction errors. 

Along with the attack,~\cite{biggio2013evasion} proposes an adversary model stating what an attacker knows about the system, how they can interact with it, etc.
This model has been created for general purpose, however, in MOTIV, we have a complete control over the system.
Furthermore, we do not want to attack per se, we rather want to improve the classifier using adversarial attacks.
Thus, we conduct evasion attacks under perfect knowledge of the system (\ie the learning algorithm, used training set and feature representation).

\textbf{Using evasion attack on a video generator.} Coming back at our video generator example presented
, we apply AdvML and look at "adversarial" videos that are produced.
To do so, we reused video configurations and labels presented in previous work~\cite{temple:hal-01323446}.

The \textit{adversary's goal} is to consider one configuration at a time and manipulates it (\ie modify its feature values) such that it becomes misclassified. In particular, an adversary can target a specific class for which the number of misclassification will increase. 

We reimplemented the evasion attack presented in~\cite{biggio2013evasion} in Python using scikit-learn. 
As evasion attacks are based on gradient descent, the separation of the classifier needs to be derivable.
Decision Trees (the classifier algorithm previously used in~\cite{temple:hal-01323446}) cannot be derived so the classifier algorithm needs to be changed.
We decided to use Support Vector Machines (SVMs) instead as they have already been studied under adversarial settings~\cite{biggio2013poisoning,biggio2014security,biggio2013evasion,biggio2012poisoning}. 
Technically, MOTIV defines 80 configuration options among which some are categorical. As SVMs do not deal properly with this kind of features, we transforming them into a set of Boolean ones.
That is, for a categorical feature proposing $n$ choices, we produce $n$ Boolean features instead.

Now, we instantiate the adversary model to our case:
the adversary goal is to manipulate non-acceptable video configurations such that they become acceptable \wrt the classifier.
We have perfect knowledge over the learning system and can directly manipulate configuration options values (which acts also as feature values).
 The number of adversarial configurations to create, the number of transformations (\ie iterations) and their "speed" are parameters of the adversarial technique.
The speed refers to the amount of displacement allowed at each iteration. It is a common issue in gradient based techniques as large displacements will get closer to the objective faster but might not converge while small displacements might lead to local optima in more iterations.

\textbf{Preliminary results.} We selected $10$ configurations randomly among the configurations that were considered non-acceptable by the oracle.
For each of these configurations, we compute the gradient of the classifier function and move towards its direction $100$ times with a displacement step equals to $0.002$.
Once those $10$ new configurations have been generated, we add them to the training set with the label of the initial configuration (\ie non-acceptable) and retrain the classifier such that it will take these new information into account.
We repeated this process of selecting $10$ configurations, moving them and retraining the classifier $100$ times.
Repeating the all process allows to select configurations added at previous step as new starting configurations to run the attack.
Even if configurations generated in previous steps are selected, different gradient directions can be followed as the training set has been modified.
 This way, new areas of the configuration space can be explored and, in the end, reducing potential errors while globally increasing the confidence given in predictions.


Figure~\ref{fig:classif_errors} exemplifies the creation of a new configuration following the evasion attack algorithm.
A configuration has been chosen and copied (\ie the star the closest to triangles is superimposed to the triangle configuration), then, based on the SVM separation, a gradient towards a low confidence area is computed
The configuration is modified iteratively such that it follows the gradient. The successive modification are represented by the solid line until the algorithm stops. The final configuration is represented by a new star lying on top of red square configurations.


\begin{figure*}
	\centering
    \begin{minipage}{.3\linewidth}
    	\subfloat[]{
        	\includegraphics[scale=0.105]{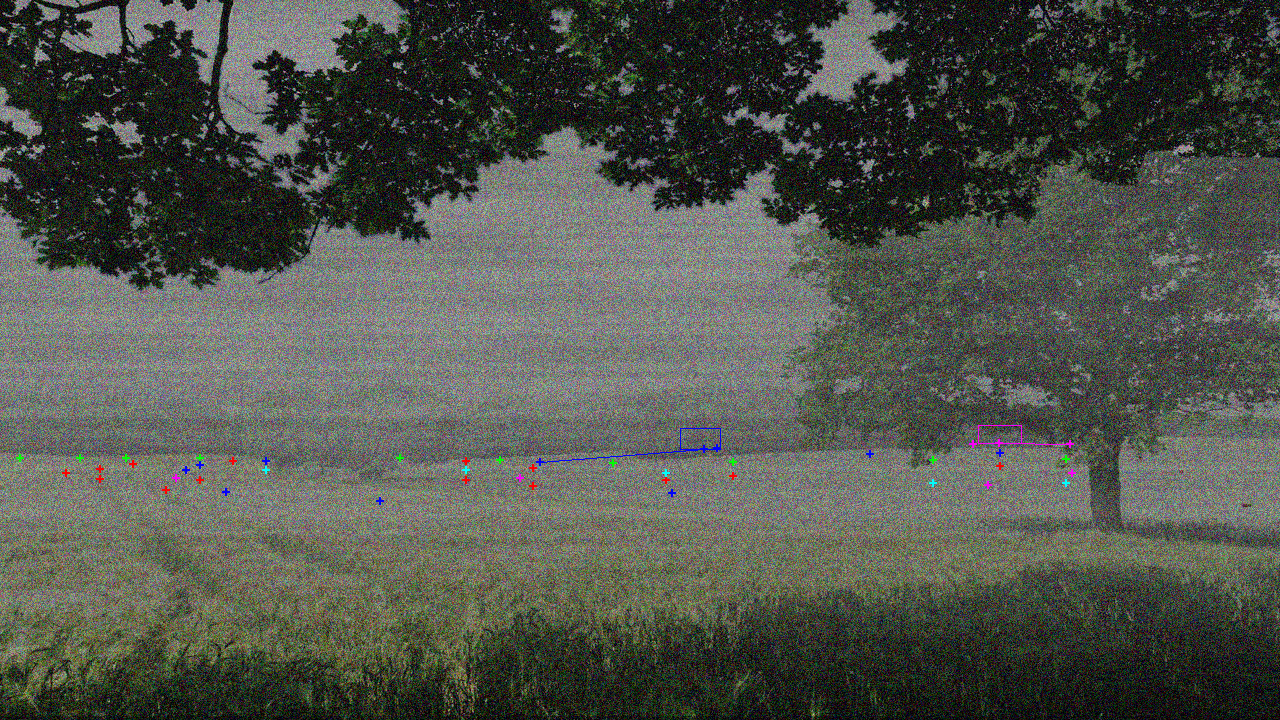}
        	\label{img:ex1}
        }
    \end{minipage}
    \begin{minipage}{.3\linewidth}
    	\subfloat[]{
        	\includegraphics[scale=0.105]{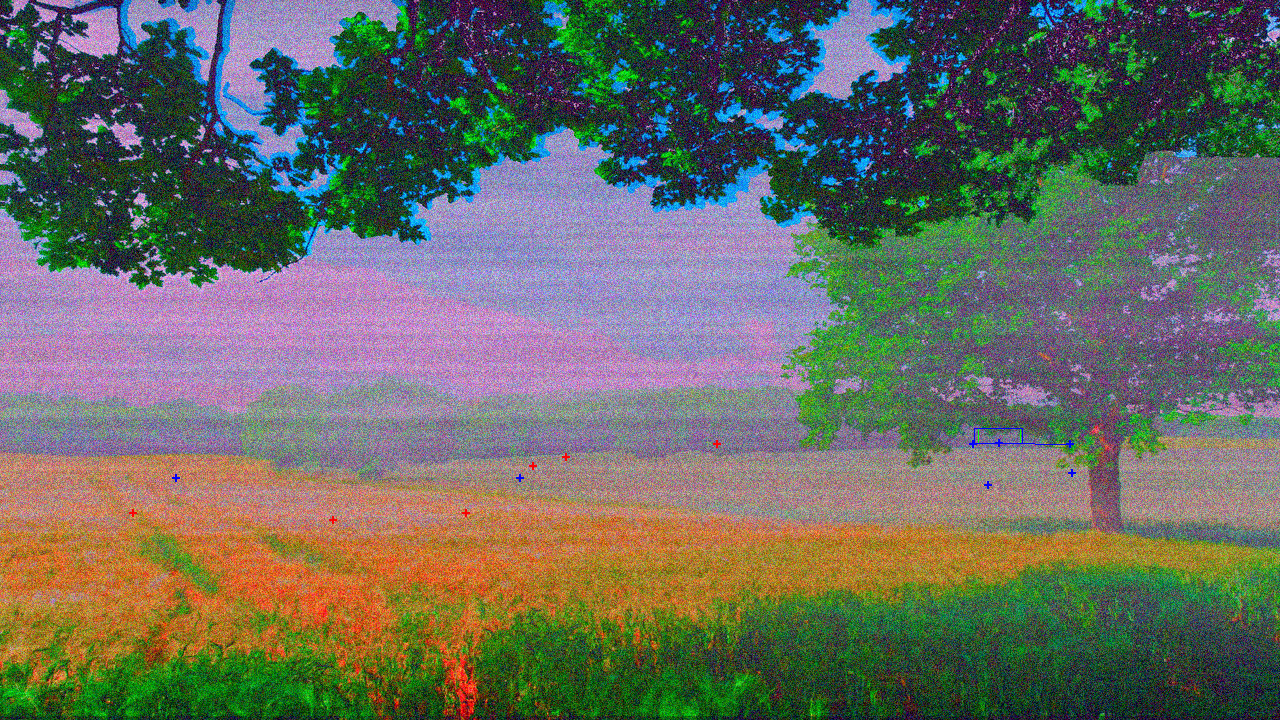}
            \label{img:ex2}
        }
    \end{minipage}
    \begin{minipage}{.3\linewidth}
    	\subfloat[]{
        	\includegraphics[scale=0.105]{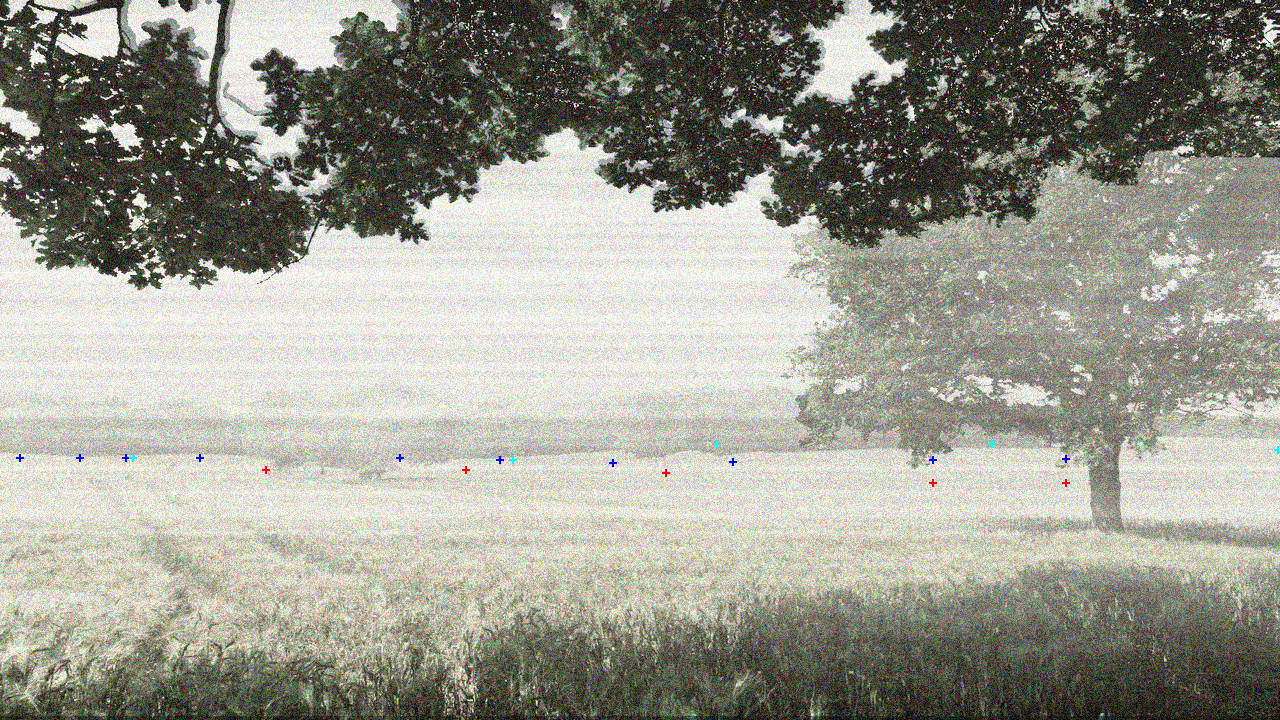}
            \label{img:ex3}
        }
    \end{minipage}
    \begin{minipage}{.3\linewidth}
    	\subfloat[]{
        	\includegraphics[scale=0.105]{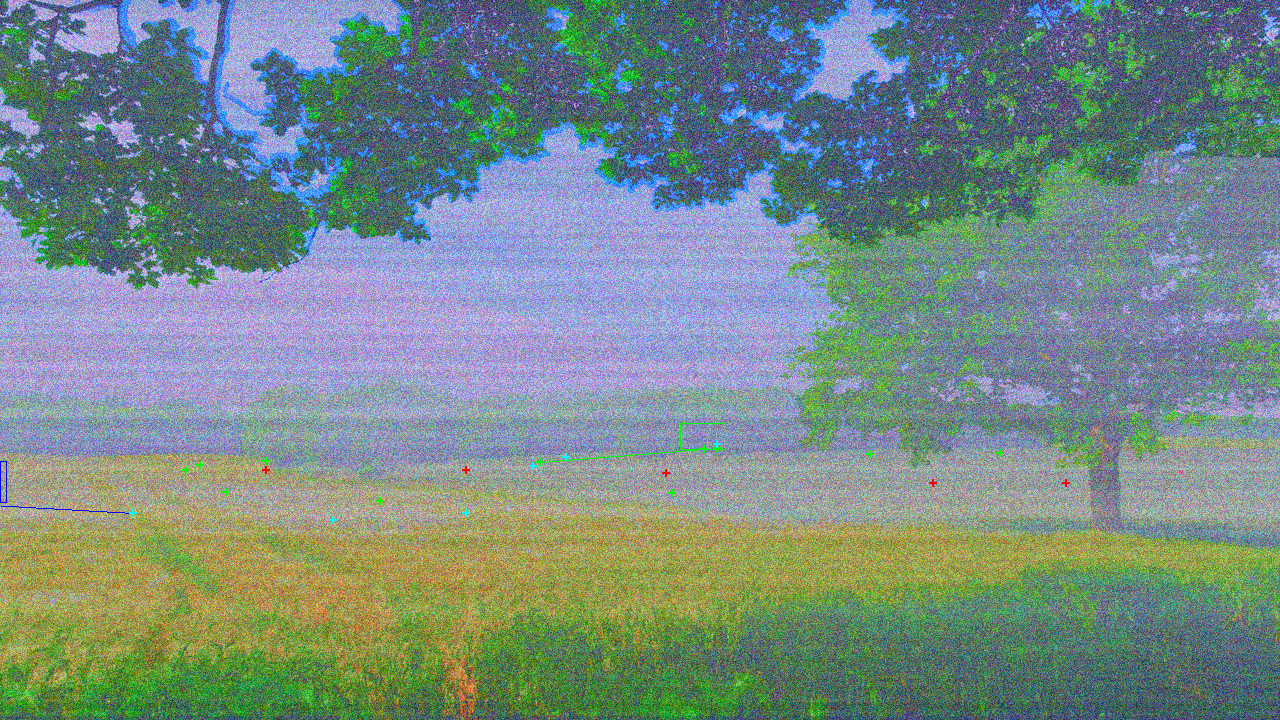}
            \label{img:ex4}
    	}
    \end{minipage}
    \begin{minipage}{.3\linewidth}
    	\subfloat[]{
        	\includegraphics[scale=0.105]{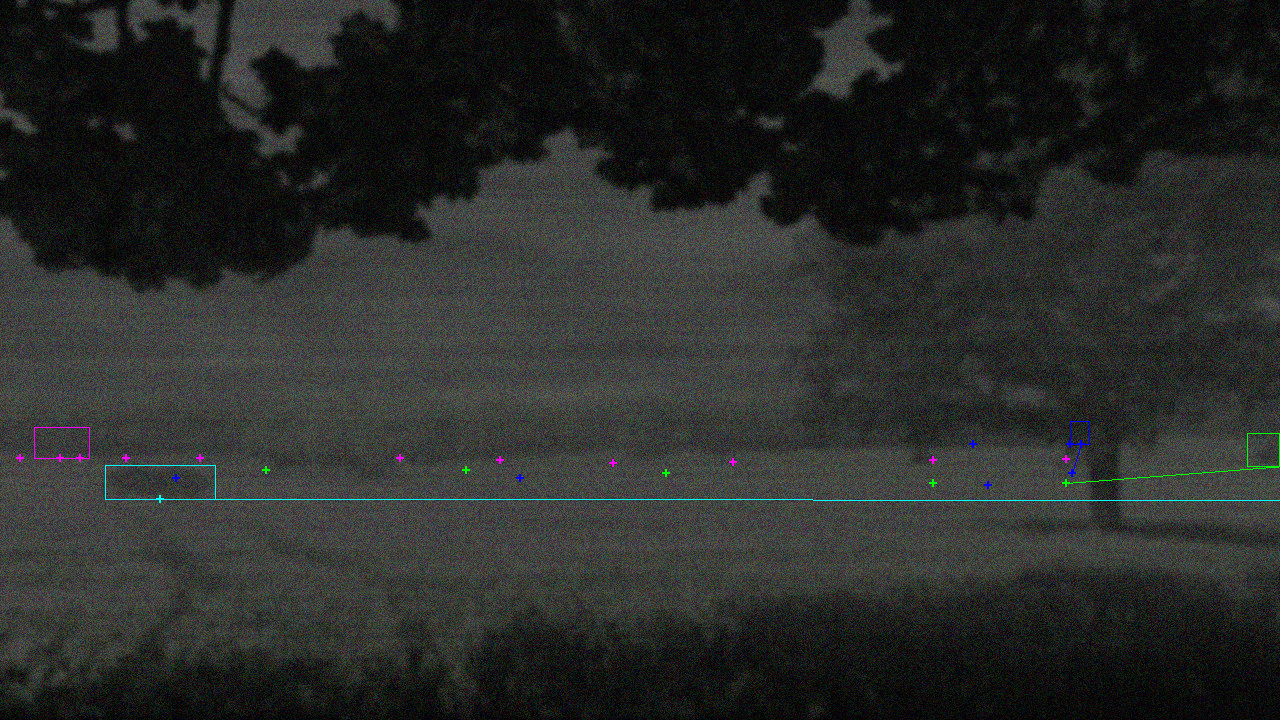}
            \label{img:ex6}
        }
    \end{minipage}
    \vspace*{-4mm}
    \caption{Examples of generated videos using evasion attack (a PDF reader is needed to appreciate the visual properties)}
    \label{img:examples_gen_vid}
\end{figure*}

Figure~\ref{img:examples_gen_vid} presents five images from five different adversarial videos that have been generated using evasion attack.
We can notice that these videos have different visual properties which means that the attack is able to consider and leverage several features.
Figure~\ref{img:ex1} shows an image where there is fog, dynamic noise and the sky is very cloudy inducing less light and thus less contrast.
The combination of fog, increasing the difficulty to identify objects in the background, and dynamic noise, a noise modifying different pixels in each frame of the video, makes it difficult to extract moving objects properly and, thus, to track them properly.
On the other hand, Figure~\ref{img:ex2} uses colors to make it more difficult for techniques matching colors to compare moving objects and models.
In particular, this image shows unrealistic colors (the variability model did not constrained the color distribution).
Figure~\ref{img:ex3} uses a heavy fog and over-exposure in order to reduce contrast and make it difficult to distinguish anything in the background.
Figure~\ref{img:ex4} uses a combination of different specific properties presented in previous images.
Figure~\ref{img:ex6} changes illumination conditions to make it looks like dark night.
As illumination is poor, the whole image can be compressed such that large homogeneous areas appear without introducing much errors in the decompression step.
In the end, large blurred areas combined to poor illumination reduce contrasts and thus makes it more difficult to detour objects and recognize them. In addition, dynamic noise (which is superimposed to the image, hence not compressed) makes it even more difficult to distinguish anything, even close to the camera.

\textbf{Capitalizing on adversarial videos.}
Our preliminary results show that evasion attack is able to combine values such that configurations will lie in new, unexplored areas.
Because, they have not been explored, they cannot be constrained properly, resulting in poor confidence in the prediction of the classifier. 
A first attitude for developers of the video generator is to include those new configurations into the training set and re-train the classifier.
Then, and following this idea presented in~\cite{temple:hal-01323446}, new rules for classification can be extracted out of the new classifier and can be added as constraints into the variability model. It will forbid to select configurations that are likely to turn into non-acceptable variants.
A second possible exploitation of adversarial configurations is to engineer a better testing oracle. Our manual review of videos indeed shows the inability of our previous automated procedure to handle such cases.  
 In particular, we can try to break down the decision process of our testing oracle into a combination of simpler procedures: a first one dedicated to assess noise level, a second one regarding blur, an other one focused on color distribution, \etc 
 Overall, adversarial configurations are helpful to improve the variability model and the testing oracle. Finally, adversarial configurations can highlight issues with variability implementation. However, in our experience, the Lua code of MOTIV was not in question.

\section{Related Work}
\label{sec:related_work}

Our work aims to support quality assurance of SPLs through the use of ML techniques. 
 Our contribution is at the crossroad of (adversarial) ML, constraint mining, variability modeling, and testing.


\textbf{Use of ML and SPL.} It has been observed that testing all configurations of an SPL is most of time impossible, due to the exponential number of configurations. ML techniques have been developed to ultimately reduce cost, time and energy of deriving and testing new configurations using inference mechanisms.
For instance, authors in~\cite{SGKA:ESECFSE15,guo2015,guo2013,DBLP:conf/isola/BeekFGS16,siegmund2013,DBLP:conf/sigsoft/OhBMS17} used regression models to perform performance prediction of configurations that have not been generated yet.
%
In~\cite{temple:hal-01323446}, we proposed to use supervised ML to discover and retrieve constraints that were not originally expressed before in a variability model. 
 We used decision trees to create a boundary between the configurations that should be discarded and the ones that are allowed. 
 In this work, we build upon this work and follow a new research direction with SVM-based adversarial learning.  
%
 Siegmund \etal ~\cite{Siegmund:2017:AVM:3106237.3106251} perform a review of ML approaches on variability models. They propose THOR, a tool for synthesizing realistic attributed variability models. An important issue in this line of research is to assess the robustness of ML on variability models. Our work specifically aims to improve ML classifiers of SPL.

In these bodies of work, none of them use adversarial ML neither the possible impact that adversarial configurations could have on the predictions.
 Our method introduces the use of evasion techniques (that is a specific adversarial ML technique) in order to specifically create configurations for "fooling" ML predictions.
Such configurations could be used in order to reinforce ML boundaries and thus give more confidence in ML predictions. 
We also show the possible impacts of adversarial configurations over variability implementation and testing oracle (not only the variability model).

\textbf{Adversarial ML} is closely related to ML as it tries to better understand flaws and weaknesses of ML techniques.
Adversarial ML can be seen as a field performing a security analysis of ML techniques.
This field has known great advances since the early 2000's with the breakthroughs of ML techniques in various domains.
Typical scenarios in which adversarial learning is used are: network traffic monitoring, spam filtering, malware detection~\cite{barreno2006can,biggio2013poisoning,biggio2014security,biggio2014pattern,biggio2013evasion,biggio2012poisoning} and more recently autonomous cars and object recognitions~\cite{deeproad2018,deepXplore2017,canFoolBothGoodfellow2018,limitDLPapernot2016,accessorize2016,advInPhysical2016,physWorldAttacks2017}.
In such works, authors suppose that a system uses ML in order to perform a classification task (\eg differentiate emails as spams and non-spams) and some malicious people try to fool such classification system.
These attackers can have knowledge on the system such as the dataset that have been used to train the ML classifier, the kind of ML technique that is used, the description of data, etc.
Based on that, they plan an attack which consists in crafting a data point in the description space that the system will mis-classify. 
%
 Recent works \cite{gan2014} have proposed to use adversarial techniques to strengthen the classifier by specifically creating data that would induce such kind of misclassification.
In this paper, we propose to use a similar approach.
However, SPLs do not suffer from an adversarial context per se. The use of adversarial techniques is rather to strengthen the SPL (including variability model, implementation and testing oracle over products) while analyzing a small set of configurations. 
%
 To our knowledge, no adversarial techniques have been used in the context of SPL or variability-intensive systems.



    

\section{Conclusion}

Machine learning techniques are more and more used in SPL engineering as they are able to predict whether a configuration (and its associated program variant) might be acceptable to end-users and their requirements.
 These techniques are based on statistical properties which can lead to prediction errors in areas where the confidence in the classification is low.
We propose to bring Adversarial Machine Learning techniques in the balance.
Exploiting knowledge over a previously trained classifier, they are able to produce so-called adversarial configurations (\ie even before variants are created) in order to specifically target low confidence areas. 
 These techniques can help detecting bugs in the mapping between configurations and actual program variants or discover additional (missing) constraints in variability models. 
  Our preliminary experiments on an industrial video generator showed promising results.
 As future work, we plan to compare adversarial learning with traditional learning or sampling techniques (\eg random, t-wise). Another research direction is to use adversarial learning for SPL regression (instead of classification) problem. In general, we want to apply the idea of generating adversarial configurations to SPLs that have large and complex configuration spaces.


\newpage
\bibliographystyle{ACM-Reference-Format}
\bibliography{./Includes/references,./Includes/macher}

\end{document}